\documentclass[11pt,a4paper]{article}
\usepackage[hyperref]{acl2019}
\usepackage{times}
\usepackage{latexsym}
\usepackage{enumitem}
\setlist[itemize]{noitemsep, nolistsep}
\usepackage{float}
\usepackage{multirow}
\usepackage{multicol}
\usepackage{graphicx}
\usepackage[justification=centering]{caption}
\usepackage{url}
\usepackage{lscape}
\usepackage{amsmath}
\usepackage{xcolor}

\aclfinalcopy 

\newcommand\Mark[1]{\textsuperscript#1}

\title{Development of POS tagger for English-Bengali Code-Mixed data}

\author{Tathagata Raha\Mark{1},
Sainik Kumar Mahata\Mark{2}, Dipankar Das\Mark{3}, Sivaji Bandyopadhyay\Mark{4}\\
\Mark{1}IIIT, Hyderabad\\
\Mark{2}\Mark{3}\Mark{4}Jadavpur University, Kolkata\\
\Mark{1}tathagata.raha@research.iiit.ac.in,
\Mark{2}sainik.mahata@gmail.com,\\
\Mark{3}dipankar.dipnil2005@gmail.com,
\Mark{4}sivaji\_cse\_ju@yahoo.com}

\date{}
\begin{document}
\maketitle
\begin{abstract}
Code-mixed texts are widespread nowadays due to the advent of social media. Since these texts combine two languages to formulate a sentence, it gives rise to various research problems related to Natural Language Processing. In this paper, we try to excavate one such problem, namely, Parts of Speech tagging of code-mixed texts. We have built a system that can POS tag English-Bengali code-mixed data where the Bengali words were written in Roman script. Our approach initially involves the collection and cleaning of English-Bengali code-mixed tweets. These tweets were used as a development dataset for building our system. The proposed system is a modular approach that starts by tagging individual tokens with their respective languages and then passes them to different POS taggers, designed for different languages (English and Bengali, in our case). Tags given by the two systems are later joined together and the final result is then mapped to a universal POS tag set. Our system was checked using 100 manually POS tagged code-mixed sentences and it returned an accuracy of 75.29\%.
\end{abstract}

\section{Introduction}
\label{sec:intro}
A \textbf{P}arts-\textbf{o}f-\textbf{S}peech (POS) Tagger is a piece of software that reads the text in some language and assigns parts of speech tags, such as noun, verb, adjective, etc., to each word/token. POS Tags are useful for building parse trees, which may be used to build textbf{N}amed \textbf{E}ntity \textbf{R}ecognizers (NER) or Dependency Parsers. POS Tagging is also useful for building lemmatizers, which are used to reduce a word to its root form. POS taggers for widely spoken languages have been developed in abundance. But such resources are very scarce for low resourced languages.

On the other hand, code-mixing is simply a mix of two or more languages in communication. Due to the emergence of social media, a lavish amount of digital code-mixed data is generated. This is because people nowadays are very comfortable with multilingualism. This phenomenon has produced a section of researchers, who contemplate code-mixed texts as being a new language. 

As mentioned earlier, since POS tagging systems for low resourced languages are hard to come by, developing one that will cater to code-mixed text is trivial. POS tagging systems, if developed for Code-Mixed data, can lead to deciphering many complex \textbf{N}atural \textbf{L}anguage \textbf{P}rocessing (NLP) tasks and hence, we attempt to develop the same in this reported work. We try to focus on creating a POS tagger for English-Bengali code-mixed data, as languages such as Bengali are morphologically rich in nature.

\begin{figure*}[h]
\centering
\includegraphics[width=\textwidth]{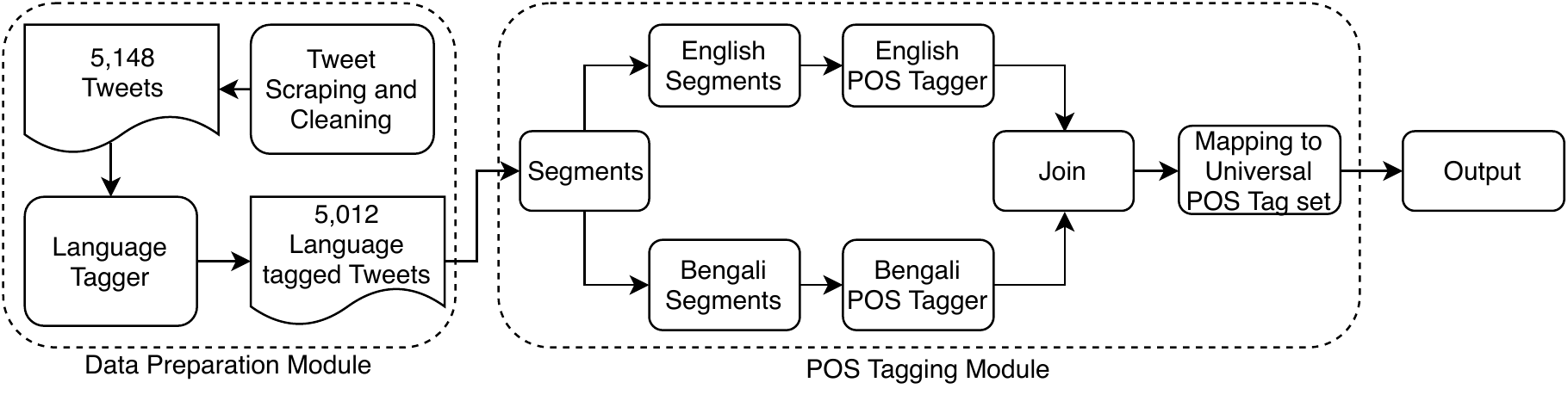}
\vspace{-1em}
\caption{Architecture of the proposed model.}
\label{fig:POS Tag model}
\end{figure*}

Our method includes scraping of code-mixed English-Bengali tweets on Twitter and cleaning them. The Bengali words in these tweets were in Roman script. These cleaned tweets were used as a development dataset for building our system. Our system starts with tagging individual tokens of a tweet with their respective languages, either English, Bengali or Unknown. This step will give rise to segments/sub-sequences of the tweet, written in the same language. It is to be noted that tokens tagged as Unknown were discarded. The segments will then be passed to two POS taggers, one designed for English and the other designed for Bengali. The output from the POS taggers will then be joined together to get the final POS tagged, code-mixed tweet. Since the POS tagging modules of English and Bengali use different tag sets, we further map the tags to a manually defined universal POS tag set. This step produces a final POS tagged tweet with uniform tags. The architecture of the proposed model is shown in Figure \ref{fig:POS Tag model}.

The remainder of the paper is organized as follows. Section \ref{sec:related_work} documents a brief state-of-art on this domain. Section \ref{sec:data_prep} defines the data preparation steps. Section \ref{sec:POS Tagging} defines the pipeline which helps us in POS tagging the code-mixed tweets. This will be followed by the results in Section \ref{sec:results} and concluding remarks in Section \ref{sec:conclusion}.

\section{Related Work}
\label{sec:related_work}
In the past few years, a lot of significant work has been done in the field of Parts of Speech tagging. The first significant POS tagger came in the early Nineties which was a rule-based tagger \cite{karlsson2011constraint}. One of the English rule-based taggers had an accuracy of 99.5\% \cite{samuelsson1997comparing}. POS taggers based on statistical approaches were also used during this time, which was based on statistical models like bi-gram,tri-gram and Markov Models \cite{derose1988grammatical,cutting1992practical,dermatas1995automatic,meteer1991post,merialdo1994tagging}. Subsequently, POS tagger based on both statistical methods and a rule-based approach was proposed by \citet{brill1992simple}.

Use of Conditional Random Fields for the development of POS taggers was proposed by \citet{lafferty2001conditional}, \citet{shrivastav2006conditional} and \citet{sha2003shallow}. \citet{nakamura1990neural} used neural networks for POS tagging for the first time.

POS taggers for the Bengali language was also built by \citet{seddiqui2003parts}. This POStagger was built on the analysis of the Bengali morphemes. Other works have been done in Bengali POS tagging by \citet{hasan2007comparison} and \citet{dandapat2007automatic} which were rule-based and semi-supervised.

\citet{pimpale2016experiments} attempted to tag code-mixed data using Stanford POS tagger. He trained the POS tagger on constrained data of Hindi, Bengali, and Telugu, mixed with English. They garnered accuracy figures of 71\%. Similarly, \citet{sarkar2016part} used the HMM model on constrained code-mixed data and achieved an accuracy figure of 75.60\%.

Pipeline architecture for POS tagging of code-mixed data was first used by \citet{barman2016part}. The training data was very low in their case and the LID (language identification) and transliteration models used were based on Support Vector Machines (SVM) and manual transliteration. Our approach also used pipeline architecture similar to theirs, but our model does not require any annotated data to train the system. Also, the LID and transliteration modules, in our case, have been fully trained with much larger data, using Deep Learning architecture.

\section{Data Preparation}
\label{sec:data_prep}
We decided to use a development dataset for building our system. It is to be noted that this data was used to build the proposed system and not to train it. Since code-mixed data consisting of English and Bengali language are difficult to find, we decided to scrape such data from Twitter. The collected tweets contained multiple degrees of noise and hence, it needed to be cleaned before using it to develop our future systems. After cleaning the tweets, they were subjected to a Language Tagger module that tagged every token of the tweet with their corresponding language (English, Bengali, and Unknown, in this case).

\subsection{Tweet Scraping and Cleaning}
\label{subsec:tweet_scrape and cleaning}
Initially, we had to assemble the development data, consisting of English-Bengali code-mixed data, that will be used to build the POS tagger model. For this, we scraped tweets from Twitter, as it is a social media handle with a huge repository of such data. Our tweet scraper module used the Twint module\footnote{https://pypi.org/project/twint/}, a python package that helps to scrape tweets. The program was fed with a list of Bengali (Romanized) keywords that will be used to scrape the tweets. Later, the Twint object iterates the keywords and recovers tweets corresponding to the same keywords.

Using this method, 5,148 code-mixed tweets containing English and Bengali (Romanized) words were collected. The collected tweets were noisy and hence we needed to clean it beforehand to proceed. The cleaning module was a manifold approach that involved cleaning links, smileys, Emojis, Hashtags, and Mentions (Usernames).

\subsection{Language Tagging and Segmentation}
\label{subsec:language_tagger}
We observed that there is no end-to-end POS tagger available that can jointly tag English and Bengali tokens. Thus we decided to segment the cleaned tweets, into Bengali and English. This was done so that tokens in different language segments can be tagged with their respective POS tags, separately.

For segmenting the tweets, the words needed to be tagged with their corresponding language. To develop such a \textbf{L}anguage \textbf{T}agging (LT) model, we collected 11,060 Romanized words of Bengali and 7,223 words of English. We developed a binary classification model that takes as input, the tokens of a tweet (in character embedding) and outputs the language of the word to either English or Bengali. Tokens (in character embedding) were fed to a stacked LSTM of size 2. The output vectors from the LSTM cells were then fed to a fully connected layer, which then mapped the words to its specific language. For the given model, Activation was kept as \textit{Sigmoid}, Optimizer used was \textit{Adam} and Loss used was \textit{Binary Crossentropy}. Batch Size was kept at 30. The program was executed for 30 epochs and the model was validated using a validation split of 0.2.

\begin{figure}[h]
\centering
\includegraphics[]{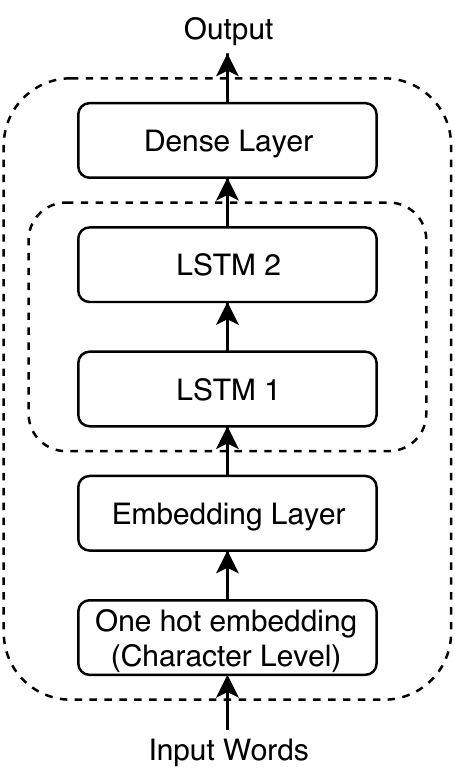}
\vspace{-1em}
\caption{Language Tagging Module.}
\label{fig:language tagging}
\end{figure}

The architecture of the language tagging module is shown in Figure \ref{fig:language tagging}. The model returned a validation accuracy of 91\%. It is to be noted that, characters apart from alphabets and numbers were tagged as ‘Unknown'. Tweets with no language tag and only unknown tags were discarded. An example of language tagging is shown in Table \ref{tab:lang_tag}. Statistics of the tweets after cleaning and language tagging are shown in Table \ref{tab:table1}.

\begin{table}[h]
\centering
\begin{tabular}{|c|}
\hline
\begin{tabular}[c]{@{}c@{}}I loved the golpo and khabar\\ ta khub nice chilo .\end{tabular} \\ \hline \hline
\begin{tabular}[c]{@{}c@{}}I\textbackslash{}en loved\textbackslash{}en the\textbackslash{}en golpo\textbackslash{}bn and\textbackslash{}en \\ khabar\textbackslash{}bn ta\textbackslash{}bn khub\textbackslash{}bn nice\textbackslash{}en chilo\textbackslash{}bn .\textbackslash{}un\end{tabular} \\ \hline
\end{tabular}
\vspace{-1em}
\caption{Example of Language Tagging.}
\vspace{-1em}
\label{tab:lang_tag}
\end{table}

\begin{table}[h]
\small
\centering
\begin{tabular}{|c|c|}
\hline
\textbf{Particulars} & \textbf{Number} \\ \hline
No. of tweets before LT & 5,148 \\ \hline
No. of tweets after LT & 5,012 \\ \hline
No. of tokens before LT & 1,44,17 \\ \hline
No. of tokens after LT & 1,41,47 \\ \hline
No. of tweets with no language tag & 136 \\ \hline
\end{tabular}
\vspace{-1em}
\caption{Statistics of tweets after cleaning and Language Tagging.}
\label{tab:table1}
\end{table}

After the language tagging is done, a segmentation module partitions the code-mixed input into segments concerning its language tags. In our case, segments are sub-sequences of the instance, written in the same language. An example of segmentation is shown below, where strings in brackets denote segments;

\begin{flushleft}
\textbf{1.} (Movie)\textsubscript{En} (ta bhalo chilo)\textsubscript{Bn} (but mid point)\textsubscript{En} (e amar khub)\textsubscript{Bn} (boring)\textsubscript{En} (lagte shuru korlo)\textsubscript{Bn}. \\
\textbf{2.} (I had to go)\textsubscript{En} (karon o khub)\textsubscript{Bn} (urgently)\textsubscript{En} (daklo amaye)\textsubscript{Bn}.
\end{flushleft}

\subsection{Language Switch Analysis}
\label{subsec:bigram_probability}
Language tagged tweets were then analyzed to examine switching patterns. For this, the tweets were tokenized and a list of bigrams was extracted. Since the tokens of the tweets are tagged with their specific language, we could find out the count of bigrams with respect to EN-EN (both tokens of biagram are in English), BN-BN (both tokens of biagram are in Bengali), EN-BN (fist token of biagram is in English and second in Bengali) and BN-EN (fist token of biagram is in English and second in Bengali).
\begin{table}[h]
\small
\centering
\begin{tabular}{|c|c|c|c|}
\hline
\textbf{Switch} & \textbf{Count} & \textbf{Freq \textgreater 500} & \textbf{Freq \textgreater 1000} \\ \hline
EN-BN & 17,758 & 199 & 88 \\ \hline
BN-EN & 17,562 & 166 & 53 \\ \hline
EN-EN & 43,859 & 539 & 203 \\ \hline
BN-BN & 16,535 & 98 & 39 \\ \hline
\end{tabular}
\vspace{-1em}
\caption{Language Switch Analysis.}
\label{tab:bigram probablity}
\end{table}

\section{Parts of Speech Tagging}
\label{sec:POS Tagging}
After the data preparation step, the language tagged segments are passed to the corresponding language POS tagger for the final tagging. Two different POS tagging systems were used for English and Bengali. For POS tagging the English Segments we used the Stanford POS tagger\footnote{https://nlp.stanford.edu/software/tagger.shtml} and the output was recorded.

For the Bengali segments, we used a tagger developed by \citet{6973488}. They trained the tagger on 10,000 Bengali (Devanagari) POS tagged sentences and tested it on 2,000 Bengali (Devanagari) sentences. Their model returned 92\% accuracy. To use their model, we had to transliterate the Bengali segments into its corresponding Devanagari script. The model developed to do the same is described in Section \ref{subsec:Bengali Transliteration}.

\subsection{Bengali Transliteration}
\label{subsec:Bengali Transliteration}
To develop the transliteration system, we initially collected 22,781 Romanized Bengali words and manually transliterated them to its Devanagari counterpart. We developed a Sequence-to-Sequence model that takes as input the Romanized Bengali words and outputs the Bengali words in the Devanagari script. The embedding used in this model was at the character level.

The model consists of two parts: an Encoder and Decoder. The encoder takes as input, Romanized Bengali characters, creates one-hot vectors of the same and passes this to the Embedding layer. The output of the embedding layer is given to a stacked LSTM cell, which produces a context vector of the input word. The Decoder module takes as input the Bengali characters in Devanagari script, creates a one-hot vector of the same and passes it to an embedding layer. The output of the embedding layer is given to a stacked LSTM cell which is initialized with the state of the encoder module. The stacked LSTM cell then produces Bengali characters (in Devanagari script) as output, with an offset of a one-time step. The activation of the model was selected as \textit{Softmax}, Optimizer used was \textit{Adam} and Loss used was \textit{Sparse Categorical Crossentropy}. Batch Size was kept at 1024. The program was executed for 50 epochs and the model was validated using a validation split of 0.1.

\begin{figure}[h]
\centering
\includegraphics[width=\columnwidth]{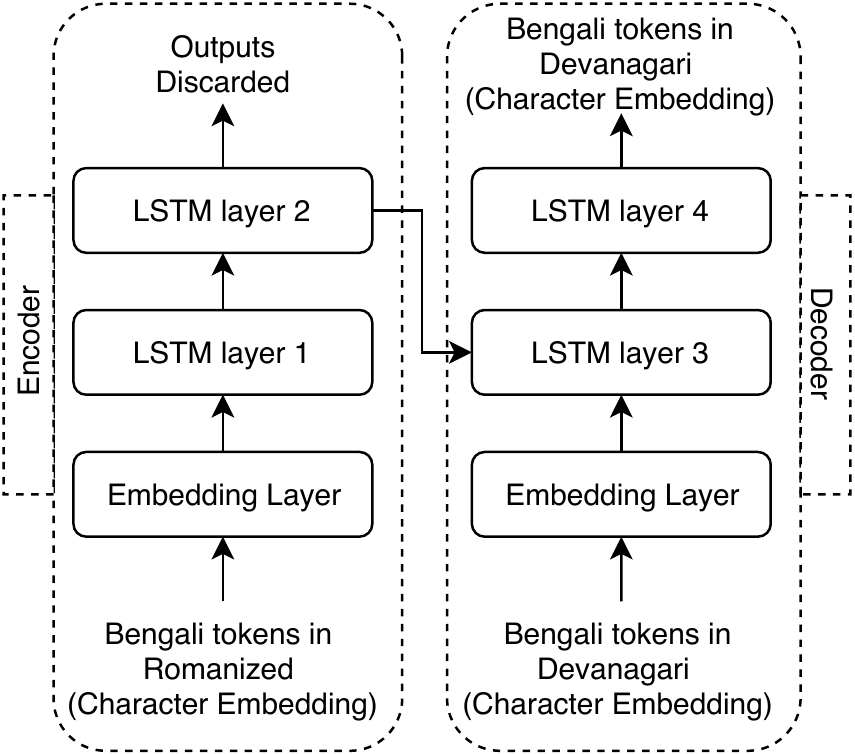}
\vspace{-1em}
\caption{Back transliteration model}
\label{fig:back_transliterate}
\end{figure}

The validation accuracy of the model was recorded as 87\%. The architecture of the model is shown in Figure \ref{fig:back_transliterate}.

The transliterated segments are then fed to the Bengali POS tagger and the corresponding outputs are recorded.

After POS tagging both the English and Bengali segments, the results are joined together to get a POS tagged code-mixed tweet.

\subsection{Mapping to Universal POS Tag Set}
\label{subsec:universal_postagset}
The final POS tagged code-mixed tweets need to be generalized to a universal system because the POS tags of the Bengali and English POS taggers are different. This is because English and Bengali POS taggers have different grammar and thus use different POS tag sets. To simplify this situation, we use a universal POS tag set that comprises the tags as showed in Table \ref{tab:universal tagset}.
\begin{table}[h]
\small
\centering
\begin{tabular}{|c|c|c|c|}
\hline
\textbf{POS} & \textbf{Univ.Tag} & \textbf{POS} & \textbf{Univ. Tag} \\ \hline
Adjective & \textit{\textbf{ADJ}} & Adposition & \textit{\textbf{ADP}} \\ \hline
Determiner & \textit{\textbf{DET}} & Noun & \textit{\textbf{NOUN}} \\ \hline
Pronoun & \textit{\textbf{PRON}} & Verb & \textit{\textbf{VERB}} \\ \hline
Adverb & \textit{\textbf{ADV}} & Conjunction & \textit{\textbf{CONJ}} \\ \hline
Numeral & \textit{\textbf{NUM}} & Particle & \textit{\textbf{PRT}} \\ \hline
Punctuation & \textit{\textbf{SYM}} & Other & \textit{\textbf{X}} \\ \hline
Demonstrative & \textit{\textbf{DEM}} & Intensifier & \textit{\textbf{INTF}} \\ \hline
\multicolumn{2}{|c|}{Reduplicative} & \multicolumn{2}{c|}{RDP} \\ \hline
\end{tabular}
\vspace{-1em}
\caption{Universal tag set, where text in bold and italics denote the tag and the text above define the tags}
\label{tab:universal tagset}
\end{table}
The table shows the universal tags in bold and italics while the other texts define the universal tag.

For mapping the English POS tags to this universal POS tag set we use \texttt{map\_tag} which is an inbuilt tool of NLTK. It maps the English tags to these tags based on some pre-defined rules.

The mapping of the Bengali POS tags (Stanford POS tags) to the universal POS tag set is shown in Table \ref{tab:Bengali_mapping}. Here, text in bold and italics denotes the universal tag, while the other defines the Stanford POS tags.

\begin{table}[h]
\small
\centering
\begin{tabular}{|c|c|c|c|}
\hline
\textbf{Syst. Tag} & \textbf{Univ. Tag} & \textbf{Syst. Tag} & \textbf{Univ. Tag} \\ \hline
NN & \multirow{3}{*}{\textit{\textbf{NOUN}}} & VM & \multirow{2}{*}{\textit{\textbf{VERB}}} \\ \cline{1-1} \cline{3-3}
NNP & & VAUX & \\ \cline{1-1} \cline{3-4}
INTJ & & JJ & \multirow{2}{*}{\textit{\textbf{ADJ}}} \\ \cline{1-3}
PRP & \multirow{2}{*}{\textit{\textbf{PRON}}} & QF & \\ \cline{1-1} \cline{3-4}
WQ & & RB & \multirow{2}{*}{\textit{\textbf{ADV}}} \\ \cline{1-3}
DEM & \textit{\textbf{DEM}} & NEG & \\ \hline
PSP & \textit{\textbf{ADP}} & RP & \textit{\textbf{PRT}} \\ \hline
CC & \textit{\textbf{CONJ}} & INTF & \textit{\textbf{INTF}} \\ \hline
QC & \textit{\textbf{NUM}} & RDP & \textit{\textbf{RDP}} \\ \hline
SYM & \textit{\textbf{SYM}} & UN & \textit{\textbf{UN}} \\ \hline
DET & \textit{\textbf{DET}} & Other & \textit{\textbf{X}} \\ \hline
\end{tabular}
\vspace{-1em}
\caption{Mapping of Bengali POS tags to the universal tagset. Text in bold and italics denotes the universal tag, while the other defines the Stanford POS tags}
\label{tab:Bengali_mapping}
\end{table}

\noindent Finally, the POS tagged segments (mapped to the universal POS tagset) are recorded as the final output.

\section{Results}
\label{sec:results}
Since there is no automated evaluation metric present to assess the quality of POS tagging a code-mixed sentence, we hired a linguist who was proficient in both Bengali and English. The linguist was asked to prepare a test data comprising of 100 English-Bengali code-mixed sentences. Further, the linguist was asked to POS tag the tokens, based on the universal POS tagset, separately. The linguist was told to look into the context of the sentence while tagging the tokens. This approach was used to properly
\begin{itemize}
\item tag ambiguous words, such as 'to', which occurs in both English and Bengali.
\item tag words in the switching point.
\end{itemize}
The same test data was tagged using our system as well. To calculate the agreement between the manual annotation and system annotation, we used Krippendorff's Alpha \cite{krippendorff2011computing}, and the metrics and the confusion are shown in Table \ref{tab:agreement_english}

\begin{table}[h]
\centering
\small
\begin{tabular}{|c|c|c|c|}
\hline
\textbf{\begin{tabular}[c]{@{}c@{}}POS\\ Tag\end{tabular}} & \textbf{\begin{tabular}[c]{@{}c@{}}Man.\\ Tag\end{tabular}} & \textbf{\begin{tabular}[c]{@{}c@{}}Syst.\\ Tag\end{tabular}} & \textbf{\begin{tabular}[c]{@{}c@{}}Diff. \&\\ Conf.\end{tabular}} \\ \hline
\textbf{NOUN} & 522 & 538 & 16 \colorbox{pink}{ADJ}\colorbox{green}{VERB} \\ \hline
\textbf{VERB} & 286 & 259 & 27 \colorbox{red}{NOUN}\colorbox{yellow}{PRON} \\ \hline
\textbf{ADJ} & 169 & 141 & 28 \colorbox{red}{NOUN}\colorbox{green}{VERB} \\ \hline
\textbf{PRON} & 104 & 118 & 14 \colorbox{pink}{ADJ}\colorbox{purple}{ADV} \\ \hline
\textbf{ADV} & 93 & 63 & 30 \colorbox{green}{VERB}\colorbox{purple}{ADV} \\ \hline
\textbf{SYM} & 59 & 60 & 1 \colorbox{orange}{NUM} \\ \hline
\textbf{CONJ} & 58 & 49 & 9 \colorbox{red}{NOUN}\colorbox{green}{VERB} \\ \hline
\textbf{DET} & 54 & 53 & 1 \colorbox{green}{VERB} \\ \hline
\textbf{ADP} & 54 & 49 & 5 \colorbox{blue}{PRT} \\ \hline
\textbf{PRT} & 21 & 18 & 3 \colorbox{pink}{ADJ} \\ \hline
\textbf{DEM} & 10 & 11 & 1 \colorbox{red}{NOUN} \\ \hline
\textbf{NUM} & 9 & 9 & 0 \\ \hline
\textbf{INTF} & 3 & 6 & 3 \colorbox{green}{VERB} \\ \hline
\textbf{RDP} & 1 & 1 & 0 \\ \hline
\textbf{UN} & 0 & 606 & 606\\ \hline
\textbf{\begin{tabular}[c]{@{}c@{}}K's \textbackslash{}alpha\\ (Interval)\end{tabular}} & \multicolumn{3}{c|}{0.7522} \\ \hline
\end{tabular}
\vspace{-1em}
\caption{Agreement Analysis between manual tagged and system tagged POS tags}
\label{tab:agreement_english}
\end{table}

Inter-system annotation agreement scores described in Table \ref{tab:agreement_english} evaluates the overall system. To dive deeper, we evaluated every sentence of the test data. This was done using two methods. \\

\noindent \textbf{Method 1:}\\
For a code-mixed sentence, the POS tag of every token in the same manually annotated sentence as compared to the POS tag of every token in the same system annotated sentence. score\textsubscript{A} was calculated as\\
\begin{equation*}
\small
\text{score\textsubscript{A}}=\frac{\text{\# matched POS tags with manual tagged sentence}}{\text{\# tokens in the manually annotated sentence}}
\end{equation*}
\\\\
\noindent \textbf{Method 2:}
POS tagging of tokens that lie in the language switching point,i.e., $word\textsubscript{English} \leftrightarrow word\textsubscript{Bengali}$, is of utmost importance as the context of the two words may change. As a result, POS tags may also differ. In this context, score\textsubscript{B} was calculated by multiplying 0.25 to score\textsubscript{A} and taking the absolute value of its $log$ value, if POS tags (for the language switching point) in the manually annotated sentence and the system annotated sentence, match. The multiplying factor was kept at 0.25 as there can be four bigrams, i.e., EN-EN, BN-BN, EN-BN, and BN-EN.

If there is more than one switching point and the POS tags match, the multiplying factor was repeated for the number of switching. So, if there are two switching points, and the POS tags match, score\textsubscript{A} will be multiplied by 0.25 and 0.25 to get score\textsubscript{B}.\\
\begin{equation*}
\small
\text{score\textsubscript{B}}=\lvert\log(\text{score\textsubscript{A}}*(0.25)\textsuperscript{n})\rvert
\end{equation*}
, where $n$ denotes the number of language switching points present and the trailing * denote that the formula holds true if certain conditions are met.

With the help of the above methods, Score\textsubscript{A} and Score\textsubscript{B} were calculated for every sentence and finally, the average for the whole test data was calculated. With method 1, our algorithm garnered accuracy of \textbf{72.72\%} and with method 2, the accuracy increased to \textbf{75.29\%}.

\section{Conclusion}
\label{sec:conclusion}
In this work, we have devised a modular system that can POS tag English-Bengali code-mixed sentences. The system uses sub-modules to perform the same. Owing to the fact, that the sub-modules can be trained for any given language, the proposed approach can be used to tag a variety of code-mixed data involving any two language pairs.

The system can be enhanced further if the sub-modules can be trained using more annotated data. E.g., if the POS tagger for the Bengali language could have been trained using more data, the problem of tagging untrained tokens with 'UN' tags could have been solved. Also, the problem of wrongly tagging tokens, e.g., tagging NOUN as ADJ, VERB and tagging PRON as NOUN, VERB, etc., could have been solved. This would have made the Bengali POS tagging module more robust. The same applies to the transliteration module as well.

In the future, we would like to develop an end-to-end system, so that the errors of one sub-module do not propagate to the other sub-modules.

\bibliography{bibliography}
\bibliographystyle{acl_natbib}

\end{document}